\documentclass[10pt, a4paper]{article}
\usepackage{lrec}
\usepackage{graphicx}
\usepackage{tabularx}
\usepackage{soul}

\usepackage{epstopdf}
\usepackage[latin1]{inputenc}

\usepackage{xstring}
\usepackage{subcaption}

\usepackage[table]{xcolor}

\newcommand{\url}[1]{\texttt{#1}}

\newcommand{\vbl}{{\vrule width 1.1pt}}
\newcommand{\hbl}{\noalign{
\hrule height 1.1pt
}}
\newcommand{\pad}{1.3}

\title{Learnings from Technological Interventions in a Low Resource Language:\\ A Case-Study on Gondi\\}

\name{Devansh Mehta\thanks{* Equal Contribution} \thanks{Correspondence: kalikab@microsoft.com}$^{1*}$, Sebastin Santy$^{2*}$\footnotemark[1], Ramaravind K.  Mothilal$^{2}$, Brij M.L. Srivastava$^{3}$, \vspace{1mm}\\\large\bfseries Alok Sharma$^{4}$, Anurag Shukla$^{5}$, Vishnu Prasad$^{1}$, Venkanna U.$^{5}$, Amit Sharma$^{2}$, Kalika Bali$^{2}$\vspace{1mm}}

\address{\large $^{1}$Voicedeck Technologies, $^{2}$Microsoft Research India, $^{3}$INRIA France, $^{4}$DN Developers, $^{5}$IIIT Raipur\\
}

\abstract{
The primary obstacle to developing technologies for low-resource languages is the lack of usable data. In this paper, we report the adoption and deployment of 4 technology-driven methods of data collection for Gondi, a low-resource vulnerable language spoken by around 2.3 million tribal people in south and central India. In the process of data collection, we also help in its revival by expanding access to information in Gondi through the creation of linguistic resources that can be used by the community, such as a dictionary, children's stories, an app with Gondi content from multiple sources and an Interactive Voice Response (IVR) based mass awareness platform. At the end of these interventions, we collected a little less than 12,000 translated words and/or sentences and identified more than 650 community members whose help can be solicited for future translation efforts. The larger goal of the project is collecting enough data in Gondi to build and deploy viable language technologies like machine translation and speech to text systems that can help take the language onto the internet. \\ \newline  }

\begin{document}

\maketitleabstract

\section{Introduction}
Around 40\% of all the languages in the world face the danger of extinction in the near future. Languages are not only a means of communication but also a carrier of tradition and cultures like verbal art, songs, narratives, rituals etc. When a language spoken in a particular community dies out, future generations lose a vital part of the culture that is necessary to completely understand it. This makes language a vulnerable aspect of cultural heritage and hence calls for their preservation. When it comes to saving such endangered languages, there are two aspects: Preservation and Revitalization (also referred to as `revival linguistics') \cite{zuckermann2013historical}. The former is concerned with how languages can be archived using different linguistic techniques so that it can serve as a lookup for future generations, while the latter focuses on ensuring that the language is resurrected into the daily fabric of people's lives. The biggest success story of a language getting revitalized is Hebrew \cite{fellman1973revival}, which went from few native speakers to several million.

Initiatives like SOAS' Endangered Languages Documentation Programme (ELDP) \footnote{SOAS ELDP: \url{https://www.eldp.net/}} and The Language Conservancy (TLC) project \footnote{TLC: \url{https://www.languageconservancy.org/}} contribute mostly towards the documentation of endangered languages. However, language evolves with culture, and focusing solely on archival efforts misses out on how societies might have evolved differently had their language continued to be in use. In the present day and age of globalization and the integration of technology into almost every aspect of life, native speakers are turning to dominant languages at a faster rate than ever before to provide greater economic and social opportunities to future generations. Any revitalization effort undertaken today needs to include technological interventions that can comprehensively reverse this degradation. At a bare minimum, these languages need to be integrated with the Internet, which is becoming an ever dominant part of our lives, to ensure their survival and continued usage.



\begin{figure}[!t]
    \centering
    \includegraphics[width=0.8\linewidth,keepaspectratio]{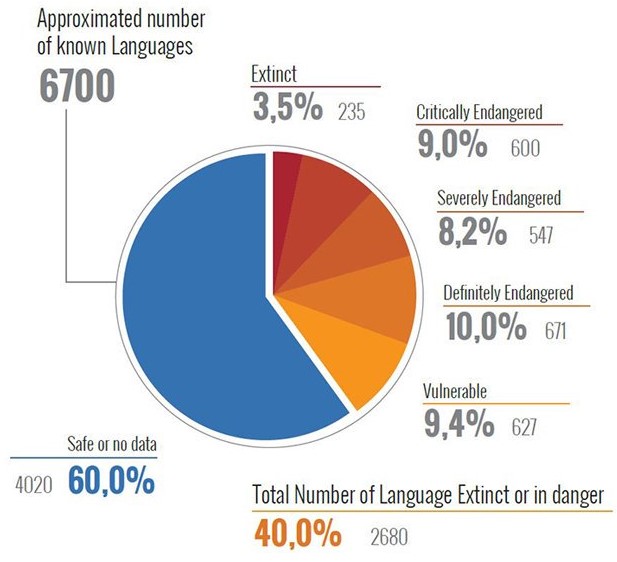}
    \caption{UNESCO 2017 World Atlas}
\end{figure}

We focus our efforts on Gondi, a South-Central Dravidian tribal language spoken by the Gond tribe in Central India. Gondi provides a unique outlook of how a language can be in danger even after having all the ingredients of a sustainable language like (1) long historical continuity (2) a population of 3 million people speaking it and, (3) is widely spoken in around 6 states of India with various dialects and forms. The complexities arise as Gondi is a predominantly spoken language with no single standard variety but a number of dialects, some mutually unintelligible. \cite{beine1994sociolinguistic}.

Deploying technology is a non-trivial task and there are legitimate concerns surrounding how language technologies should be implemented for low-resource languages \cite{joshi2019unsung}. It is difficult to simply transfer technologies prevalent in high resource language communities to minority communities for many reasons, the chief among them being the lack of data in low-resource languages. Our focus when working with the Gond community is thus centered more around devising novel approaches for data collection, unlike well-resourced languages where the focus is more on engineering. 
There also needs to be groundwork and identification of the real problems that can be solved by deployment of language technologies in minority communities. \cite{dearden2015ethical}. Technical systems working in isolation from social contexts can be very dangerous to the ecosystem of these minority language communities and hence aren't ethically neutral exchanges of information. As far as possible, we have ensured that the technology interventions described in this study were led by the Gond community and non-profit organizations with a long history of working with the community. We have a deep appreciation of the fact that the problem we are trying to tackle is different from solving an engineering problem, since the expected outcomes have social rather than technical consequences.

In this paper, we deploy 4 technological interventions to help revitalize Gondi. The interventions are designed to achieve two objectives: (1) create a repository of linguistic resources in Gondi, with the intention of eventually using them to build language technologies like machine translation or speech to text systems that are essential for taking Gondi onto the internet; and (2) expand the information available to the Gond community in their language. 

The first linguistic resource created is a Gondi dictionary that is accessible to the community as an Android app. The second is 230 children's books that were translated by the Gond community in a 10 day workshop. The third is an Android app that serves as a one-stop shop for Gondi content on the internet and also crowdsources translations from the community. The final intervention is a phone number that community members could call to gain awareness about a local election in their area, upon completion of which they earn mobile credits. These interventions respectively resulted in identifying 80-100 community members that participated in creating the 3,500 word dictionary; 20 community members that translated about 8000 sentences from Hindi to Gondi; 7 community members that translated 601 words; and 557 native speakers of Gondi that can be called for future workshops. The goal of the project is working with the Gond community members to collect enough data to build machine translation and speech to text systems that can help take Gondi onto the internet.









\section{Context}

According to the 2011 census \cite{chandramouli2011census}, the total population of the Gond tribe is approximately 11.3 million. However, the total Gondi speaking population is only around 2.7 million. That is, only about 25 percent of the entire tribe now speaks it as a first language. UNESCO's Atlas of the Worlds Languages in Danger \cite{moseley2010atlas} lists Gondi as belonging to the vulnerable category. There is an added difficulty of creating resources for Gondi due to the linguistic heterogeneity within the Gonds. Spread over 6 states in India, Gondi is heavily influenced by the dominant language of each state to the point where a Gond Adivasi from Telangana (a Telugu speaking Southern state), finds it difficult to understand a Gond Adivasi from Chhattisgarh (a Central state with Hindi as the dominant language).

\begin{figure}[!t]
    \centering
    \fcolorbox{lightgray}{white}{\includegraphics[width=0.8\linewidth,keepaspectratio]{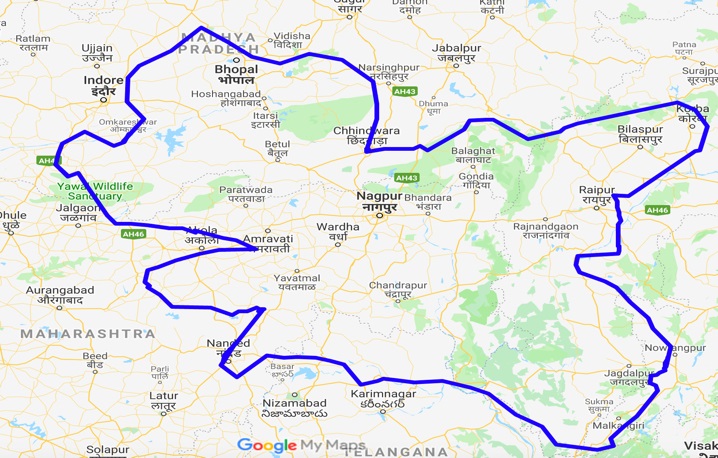}}
    \caption{Gondi speaking areas in India}
\end{figure}

As a predominantly oral language, the proportion of Gondi speakers is expected to go down further as opportunities are shrinking to hear the language spoken outside of their everyday surroundings. For example, All India Radio, the only radio station in India allowed to broadcast news, does not have a single Gondi news bulletin in their broadcasts. There is no TV station or channel catering to Gondi speakers. There is also a severe dearth of online content in Gondi, resulting in members of the tribe having to learn a mainstream language in order to enjoy the benefits of internet connectivity. One of the few exceptions to this is ``The Wire"\footnote{Gondi Bulletin: \url{https://youtu.be/M3q2ycJ\_U7g}}, an Indian news outlet that has a news bulletin in Gondi that is published on Youtube. Views of their Gondi news broadcast range from 2,500 to 7,500, although it should be kept in mind that the Gondi spoken in the broadcast caters to the Gond Adivasis from the Central states of India and it is difficult for Gonds from the Southern states to understand the content.

Gondi is also not included in the 8\textsuperscript{th} Schedule of the Indian Constitution, with the result that education and exams for government jobs cannot be administered in the language. The deleterious effects of this on their society are manifold. Gondi is considered the lingua franca of the local insurgents, who use the knowledge of the language and the perceived neglect by the government to recruit candidates from the tribe to join them \cite{kumar2019gondi}. Further, there are high dropout rates among children that speak Gondi as a first language. A 2008 UNESCO study found that children whose mother tongue is not the medium of instruction at primary school are more likely to fail in early grades or drop out, which in turn increases the chances of them joining the insurgency \cite{buhmann2008mother}. Working with the Gond tribes on reviving their language is thus important not just for cultural reasons, but may also serve as an instrument for bringing peace to their society. Reports from workshop participants attest to the younger generation in the Gond tribe attaching less importance to developing their mother tongue. However, we have also found committed community members determined to revive their language and it proved easy to recruit participants for the 8 workshops we held with the community.

\section{Technological Intervention}

There are relatively few cases that provide a holistic view of technology being used to help revive usage of an endangered or vulnerable language. The larger framework our interventions fall under is that language resource creation feeds into building language technologies, which in turn enhances access to information in that language. To give an example of how this framework might operate, ensuring that Gondi is usable and accessible online greatly enhances the access to information. However, building the necessary language technology to take it online requires copious amounts of language data - by one estimate, a minimum of 100,000 translated sentences are required to build a machine translation tool \cite{koehn-knowles-2017-six}, while at least 500 hours of transcribed and translated Gondi audio content is needed for a speech to text interface.\cite{huang2014historical}

Our efforts at collecting Gondi language resource creation and enhancing access to information in Gondi can be placed under broadly four buckets:

\subsection{Gondi Dictionary Development}

The Gondi speaking community is spread over 6 states in India. The dominant language in each state has crept into the dialect spoken by the Gond tribe residing there, with the result that it has become difficult for speakers from different parts to understand one another easily. In order to facilitate a democratized process of mutual intelligibility between the Gond groups, CGNet Swara organized seven workshops with Gondi speaking representatives from 6 states to develop a Gondi thesaurus containing all the different words used by speakers from various Gond regions. Some words, such as water, had as many as 8 different words for it. At the 8\textsuperscript{th} workshop in 2018, which saw more than 80 people in attendance, the thesaurus was developed into a dictionary containing 3,500 words that were understandable by all groups, in essence `purifying' the language of words that had entered their dialect due to the influence of the dominant state language. This dictionary not only enables translation of key words between Hindi and Gondi, but also understanding between the Gond tribes themselves. \footnote{\url{aka.ms/indian-express-gondi-dictionary}} 

To enhance reach, the dictionary was made into an Android app, Gondi Manak Shabdkosh\footnote{\url{aka.ms/gondi-dictionary}}, that allows users to enter a Hindi or Gondi word and immediately hear or read its equivalent translation. This app is in many ways similar to the Ma! Iwaidja dictionary app\footnote{\url{ma-iwaidja-dictionary.soft112.com/}}, except that there is no wheel based interface for conjugation and sentence formation. 

\begin{figure}[!t]
    \centering
        \begin{subfigure}{0.49\linewidth}
            \fcolorbox{lightgray}{white}{\includegraphics[width=\columnwidth,keepaspectratio]{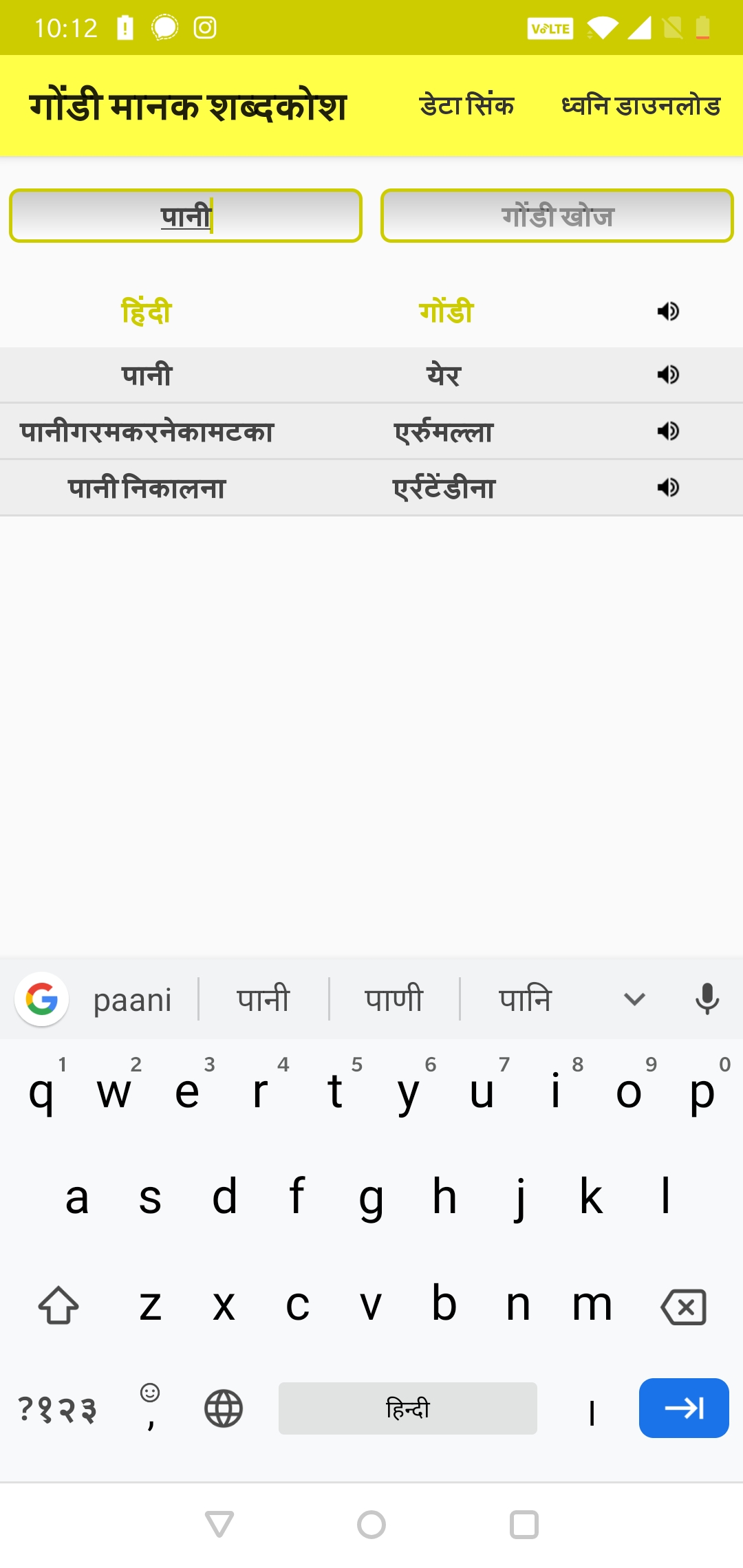}}
        \end{subfigure}\hfill
    \caption{Gondi Manak Shabdkosh}
    \label{fig:gondimanak}
\end{figure}{}

One of the use cases being explored with this app is in primary education, where tribal students and teachers are sometimes unable to communicate as the teacher does not know Gondi while the student does not know any other language. The hope is that the dictionary app will allow some basic communication and learning to take place.

\subsection{Creating children's books in Gondi}

One of the most effective ways of reviving a language is introducing it as a medium of instruction in schools, or at least as a subject. However, creating textbooks in that language is an important first step to achieving this goal.

Pratham Books is a non-profit publisher with a motto ``Book in every child's hand". Storyweaver\footnote{\url{storyweaver.org.in/}} is an initative of Pratham Books that hosts more than 15,000 children's stories in various languages and dialects. A 10-day workshop was organized where 20 bilingual Gond tribals from three states of India came together and translated children's books from Hindi to Gondi on the Storyweaver platform. These books were published on Storyweaver's website\footnote{\url{aka.ms/storyweaver-gondi}}, resulting in the first ever online repository of children's stories in Gondi.\footnote{Article: \url{aka.ms/hindu-gondi-nextgen}}

At the end of the workshop, 230 books and about 8000 sentences were translated from Hindi to Gondi. These stories, many of which introduced Gondi children to climate change for the first time, were printed out and distributed in primary schools in the Central Gondi speaking belt of Chhattisgarh. Efforts are now ongoing to convince the state government to include them as part of the school curriculum across the tribal belts of the state.
 \footnote{Article: \url{aka.ms/news18-climatechange}}

\subsection{Crowdsourcing Gondi translations}

At the Pratham Books translation workshop, it was found that many participants wanted to continue the translation work from home, but there was no avenue for them do so. Taking inspiration from Aikuma\cite{bird2014aikuma}, we developed Adivasi Radio, an Android application that presents users with Hindi words or sentences for which they need to provide the Gondi translation.

In the one month since its launch, there have been a total of 601 appearances translated through the app from 7 unique users. However, 5 of these users are either paid staff or volunteers hoping to become paid staff, indicating that soliciting translations from the community without monetary compensation may be a challenge. Moreover, the bulk of the translations, 493, have come from one superuser. Future translation workshops will include a component on continuing the translation work through the app, with the hope that attendees continue translating even after they return to their village.

In addition to the translation role, Adivasi Radio also serves as a one-stop shop for Gondi content on the internet. The translated books from Pratham are accessible through the app, as are the Gondi stories from CGnet Swara. The information on The Wire's Gondi news bulletin are also available here.

\begin{figure}[!t]
    \centering
    \begin{subfigure}{0.49\linewidth}
            \fcolorbox{lightgray}{white}{\includegraphics[width=\columnwidth,keepaspectratio]{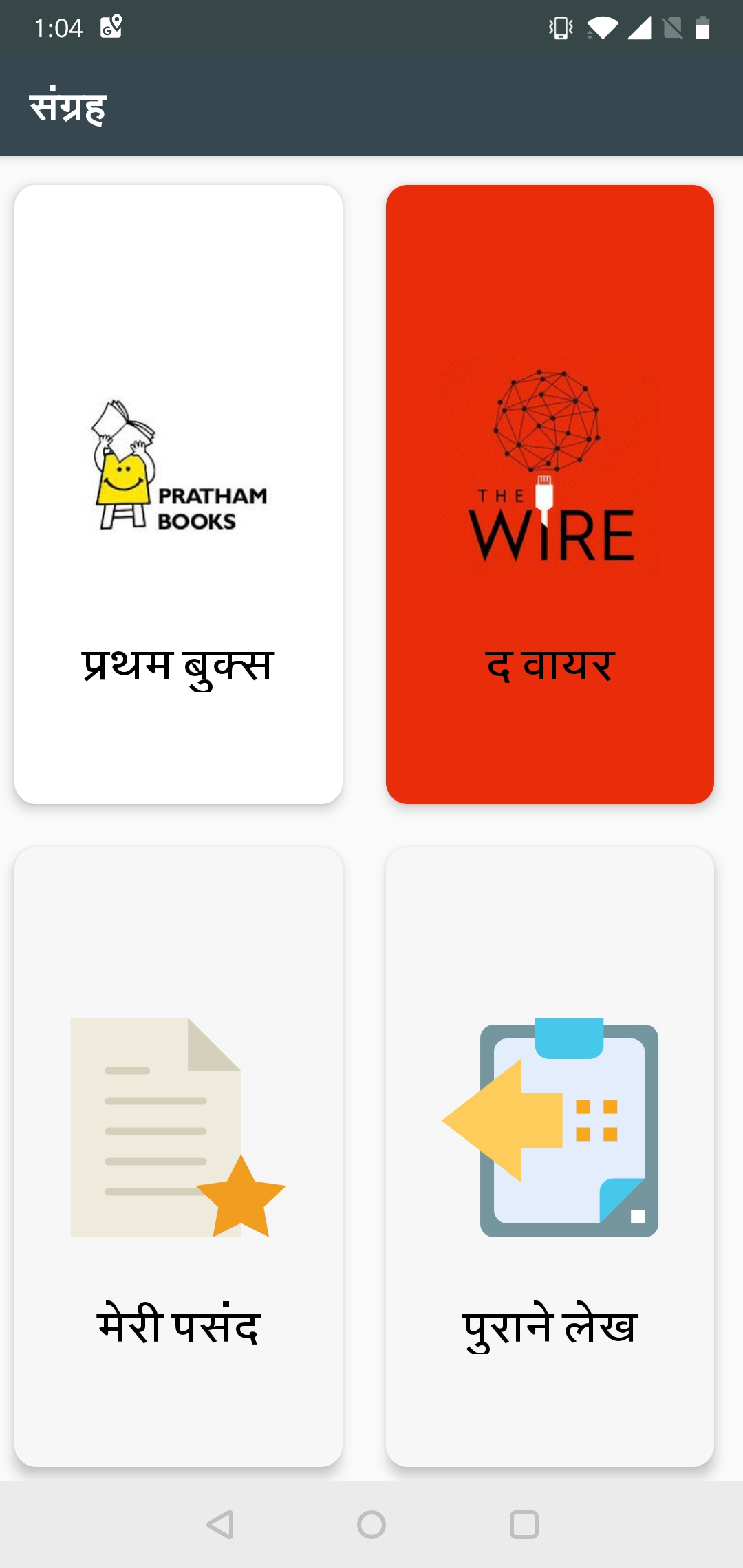}}
        \end{subfigure}\hfill
        \begin{subfigure}{0.49\linewidth}
            \fcolorbox{lightgray}{white}{\includegraphics[width=\columnwidth,keepaspectratio]{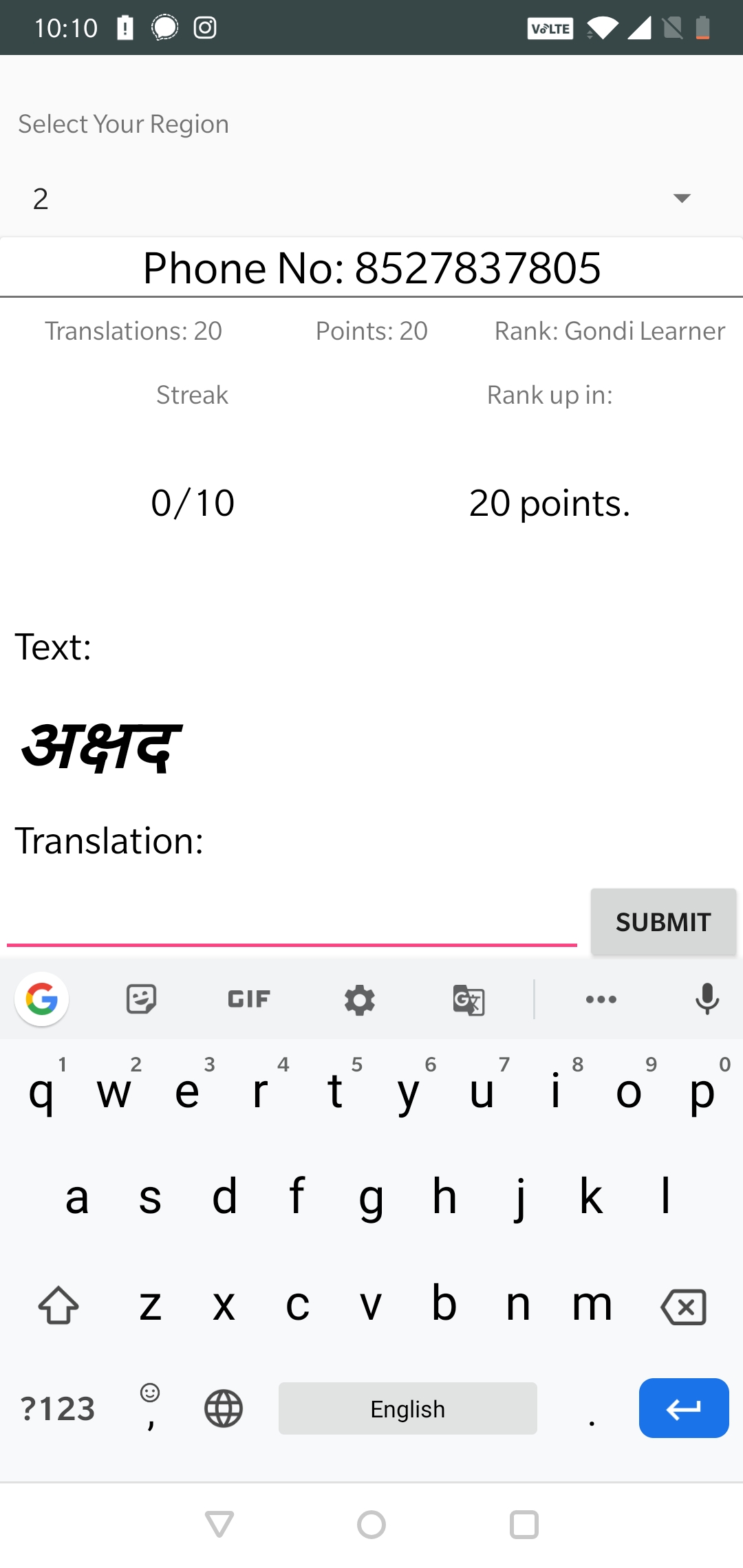}}
        \end{subfigure}\hfill
    \caption{Adivasi Radio}
    \label{fig:adivasiradio}
\end{figure}{}

To ensure these are comprehensible to populations that cannot read or write, a bootstrapped Devanagari text-to-speech system is integrated within the app to read out the written reports in Gondi. As more Gondi content proliferates online, we envisage Adivasi Radio as not only becoming the go-to place for native speakers to find outlets and sites publishing Gondi content, but also the primary medium for collecting the 100,000 sentences needed to build a machine translation tool between Gondi and other mainstream languages.

\subsection{Disseminating Gondi content}
Many recent studies on information dissemination in low-resource settings have relied on technologies, such as Interactive Voice Response (IVR), that could be used with internet-less mobile phones to connect to people and deliver information \cite{swaminathan2019learn2earn,dipanjan2019,revisitCG,designLess,ivrFarmers,raza2018baang,sawaal,raza2013job,sangeet}. As an example, Learn2Earn \cite{swaminathan2019learn2earn} is an IVR based system that was used as an awareness campaign tool to spread farmers' land rights in rural India. Learn2Earn awards mobile talktime to users who call a toll-free number and listen to an awareness message, and answer all multiple-choice questions on the message correctly. Further, it has a peer-to-peer referral component where an additional recharge is provided for every new user successfully referred to the system. Learn2Earn was successful in spreading land rights awareness to 17,000 farmers in 45 days, starting from an initial set of just 17 farmers \cite{swaminathan2019learn2earn}.

In this intervention, we adapted the Learn2Earn technology to spread voter awareness among Gondi speakers in Dantewada (a rural district in the state of Chhattisgarh in India), during the time that a bypoll election was held. We chose to disseminate voter rights messages in Gondi, as the bypoll was crucial for Gondi speakers to establish representative and effective governance in their area. Further, prior work has shown that larger the contexts, identities and communicative functions associated with the language use, the more likely the language is expected to survive. \cite{walsh2005will} We believe voter rights content in local language during elections has the potential to encourage conversations on topics of wider contexts and functions.



\begin{figure*}[!t]
    \centering
    \includegraphics[width=\linewidth,keepaspectratio]{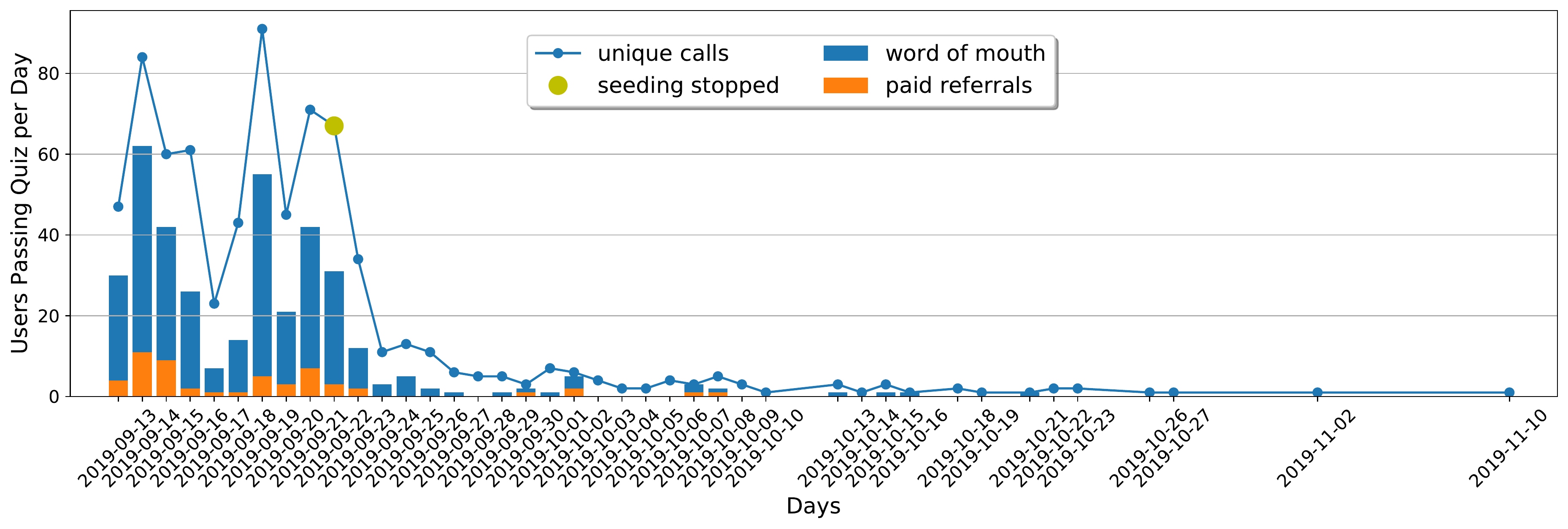}
    \caption{Number of calls to the system (blue line) and the number of users who answered all questions correctly (bar plot). The blue and orange bars represent those users who came to know about the system through word of mouth and through referrals, respectively.}
    \label{fig:usage}
\end{figure*}

\definecolor{lightgray}{gray}{0.9}
\begin{table}
\small
\rowcolors{2}{lightgray}{}
{\renewcommand{\arraystretch}{\pad}
\begin{tabular}{! \vbl p{2in}|l! \vbl}
\hbl
\cellcolor[HTML]{C0C0C0} \textbf{Metrics}                                                                                   & \cellcolor[HTML]{C0C0C0} \textbf{Value}     \\ \hline
\begin{tabular}[c]{@{}l@{}}Unique callers \\ (total, during and after seeding)\end{tabular}     & (557, 480, 77)   \\ \hline
\begin{tabular}[c]{@{}l@{}}Unique callers per day - during \\ seeding (min, mean, median, max)\end{tabular}     & (13, 48, 49, 80)   \\ \hline
\begin{tabular}[c]{@{}l@{}}Unique callers per day - after\\ seeding (min, mean, median, max)\end{tabular}     & (0, 4, 3, 20)   \\\hline
\begin{tabular}[c]{@{}l@{}}Callers answering all questions \\ correctly\end{tabular} & 313 \\ \hline
\begin{tabular}[c]{@{}l@{}}Callers answering all questions \\ correctly in their first call\end{tabular} & 104               \\ \hline

\begin{tabular}[c]{@{}l@{}}Calls made by callers\\ (min, mean, median, max)\end{tabular}          & (1, 3, 2, 64)      \\ \hbl
\end{tabular}}
 \caption{Summary statistics for our IVR-based deployment.}
\label{tab:summary}
\end{table}
Learn2Earn was previously deployed in Hindi to spread awareness for voting awareness \cite{kommiya2019learnings}. We extrapolated these experiments to Gondi to do that same in preparation for an upcoming bypoll in a Gondi speaking district. Similar to the original Learn2Earn implementation, the experiment started with a few `seed' users and is intended to spread voluntarily through word of mouth or through peer-referrals thereon, both increasing the chances of language and content spread. Figure \ref{fig:usage} shows the number of unique calls that were made to our system and the number of users who passed the awareness quiz over time. Further, the figure shows that a majority of quiz passers came to know about our system organically through word of mouth, though people knew that additional credits can be earned through referrals. Only a very few users continued to use the system even after the elections in September and after the seeding was stopped, probably to earn additional mobile talktime. 

Table \ref{tab:summary} describes important descriptive statistics of our experiment. We were able to spread voter awareness in Gondi to 557 speakers of which a majority (86\%) were reached when we actively seeded. Nearly 60\% of the users correctly answered all questions, either in their first attempt or in successive attempts, indicating the effectiveness of the system for content dissemination. Further, many users called the system more than once, with one user calling 64 times (to refer someone, users had to place another call to the system). A follow-up survey revealed that 22 of 113 users surveyed could not vote as they did not have a voter ID card, potentially useful information for authorities to increase voter turnout. One disappointing finding was that only about 8.3 percent of users were women, indicating that a focus on outreach towards this community is needed. 

We see three clear benefits from conducting Learn2Earn pilots in endangered or vulnerable languages. First, since it is entirely in the spoken form, only speakers of the endangered or vulnerable language can comprehend the content and earn the reward. Second, the phone numbers of the 557 users collected are an important dataset of speakers of that language and can be used for future translation workshops and related programs that help their economic and linguistic development. Finally, an oral language has a tendency to die out unless there is an opportunity for that language to be used outside of everyday surroundings, which periodic Learn2Earn campaigns on important issues help achieve.

\section{Prior Work}
\subsection{Preservation Programs}
The Endangered Languages Documentation Programme (ELDP) engages in documentation of endangered languages by creating a repository of resources for linguistics, the social sciences, and the language communities themselves. The Language Conservancy is a non-profit organization that strives towards revitalization of the world's endangered languages, restoring them to health and stability, and safeguarding them for future generations. They intend to implement this by introducing it as medium of instruction in schools by developing appropriate teaching methodologies and promoting and supporting the use of such languages beyond this setting - in homes and communities.

\subsection{Language Revitalization Efforts and Successes}
Hebrew is one of the languages which has shown the most promise in revitalization. Not being used since the second century, the usage of language in common conversations was kick-started by some Jewish communities in the 19$^{th}$ century \cite{fellman1973revival}. It has now risen to become the official language for the state of Israel and is spoken fluently by 7 million people.

However, such organic growth is almost impossible when the usage of language at this time is reinforced by many more factors. Ojibwe is an interesting example, where computer-based language learning technology was used and studies were conducted on how a particular multimedia tool might jumpstart communication in the Ojibwe language at home.\cite{hermes2013ojibwe} Additionally, there were efforts where twitter was seeded with many Ojibwe tweets and hashtags to motivate individuals to converse in Ojibwe on twitter.

\subsection{Deployed Applications and their Effects}
Documentation and easy access to it is the first step in reviving a language. The Ma! Iwaidja is a mobile phone app which runs a dictionary of 1500 word and 450 English-Iwaidja translations along with audio for each. In addition, it provides a functionality to add words to the dictionary by the lay user. Due to their ubiquitous nature, crowd-sourcing translations through apps are ideal due to their convenience and ease of use. Steven Bird's Aikuma app has leveraged its intuitive design to crowdsource 10 hours of audio or 100,000 words from indigenous communities in Brazil, Nepal and Papua New Guinea. \cite{bird2014aikuma} 

\section{Conclusion}

While working with well-resourced languages, the main problem in designing language technologies is engineering. For low-resource languages, however, the main problem is one of designing methods for data collection upon which the language technology can be built.

To keep community members motivated through the process of collecting the large amounts of translations needed, our team strived to achieve 2 simultaneous goals in each of the four technology interventions: the collection of data upon which language technologies such as speech to text or machine translation can be built, and expanding the access to information in that language, which the community members could point to as a demonstrable success. For example, the Learn2Earn pilot in Gondi not only provided Gond tribals with an opportunity to earn money for answering a quiz in their language and referring others to it, but it also provided a dataset of native Gondi speakers that can be called for future translation workshops. Similarly, translating children's stories and creating a standardized dictionary resulted in both data upon which machine translation tools can be built, and also tangible language resources that can be used by their community.

The Chhattisgarh government recently allowed primary education in 10 tribal languages, including Gondi \footnote{https://indianexpress.com/article/governance/chhattisgarh-education-reforms-tribal-languages-to-be-a-medium-of-education-in-pre-school-6271547/}. Our plan is to work with the government in creating new resources in Gondi that can be used in schools and distributing the translated childrens stories through the government apparatus. The larger goal of our research project is integrating Gondi with the internet, which requires at least 100,000 translated sentences; and building a speech to text platform, requiring at least 100 hours of transcribed and translated transcripts. Our most immediate goal is getting more users on the Adivasi Radio app and crowdsourcing translations through it. We plan to make all the data collected openly available and invite other researchers to participate in building language technologies that can benefit the Gond community.

\section{Acknowledgements}
We would like to thank Shubhranshu Choudhury from CGnet Swara, William Thies from Microsoft Research New England, Amna Singh from Pratham Books, and the director of IIIT Naya Raipur, Pradeep Kumar Sinha, for their invaluable assistance and advice. A word of thanks to all participants and workshop attendees from the Gond community, without whom this would not have been possible or meaningful. We are also grateful for the financial contributions made by the National Mineral Development Corporation (NMDC), South Eastern Coalfields Limited (SECL), Indira Gandhi National Center for Arts (IGNCA) and Microsoft Research India (MSR) towards supporting this research.

\section{Bibliographical References}

\bibliographystyle{lrec}
\bibliography{lrec2020W-xample-kc}


\end{document}